\pdfoutput=1
\documentclass[11pt]{article}

\usepackage[preprint]{acl}

\usepackage{times}
\usepackage{latexsym}
\usepackage[T1]{fontenc}
\usepackage[utf8]{inputenc}
\usepackage{microtype}
\usepackage{inconsolata}

\usepackage{graphicx}

\usepackage{bookmark}
\usepackage{tcolorbox}
\usepackage{tabularx}
\usepackage{booktabs}
\usepackage{multirow, multicol}
\usepackage{enumitem}

\newcolumntype{Y}{>{\centering\arraybackslash}X}

\newcommand{\ie}{{\textit{i.e., }}}

\newcommand{\rom}[1]{\lowercase\expandafter{\romannumeral #1\relax}}

\newcommand{\name}{\textsc{\textsf{RONA}}}

\title{{\name}: Pragmatically Diverse Image Captioning with Coherence Relations}

\author{
  \textbf{Aashish Anantha Ramakrishnan\textsuperscript{1}},
  \textbf{Aadarsh Anantha Ramakrishnan\textsuperscript{2}},
  \textbf{Dongwon Lee\textsuperscript{1}}
\\
  \textsuperscript{1}The Pennsylvania State University;
  \textsuperscript{2}National Institute of Technology, Tiruchirappalli
\\
  \texttt{
    \textsuperscript{1}\{aza6352, dul13\}@psu.edu, \textsuperscript{2}106121001@nitt.edu
  }
}

\begin{document}
\maketitle
\begin{abstract}
Writing Assistants (e.g., Grammarly, Microsoft Copilot) traditionally generate diverse image captions by employing syntactic and semantic variations to describe image components. However, human-written captions prioritize conveying a central message alongside visual descriptions using pragmatic cues. To enhance caption diversity, it is essential to explore alternative ways of communicating these messages in conjunction with visual content. We propose {\name}, a novel prompting strategy for Multi-modal Large Language Models (MLLM) that leverages {\em Coherence Relations} as a controllable axis for pragmatic variations. We demonstrate that {\name} generates captions with better overall {\em diversity} and ground-truth {\em alignment}, compared to MLLM baselines across multiple domains. Our code is available at: \url{https://github.com/aashish2000/RONA}
\end{abstract}

\begin{figure*}[tb]
    \centering
    \includegraphics[width=\linewidth]{"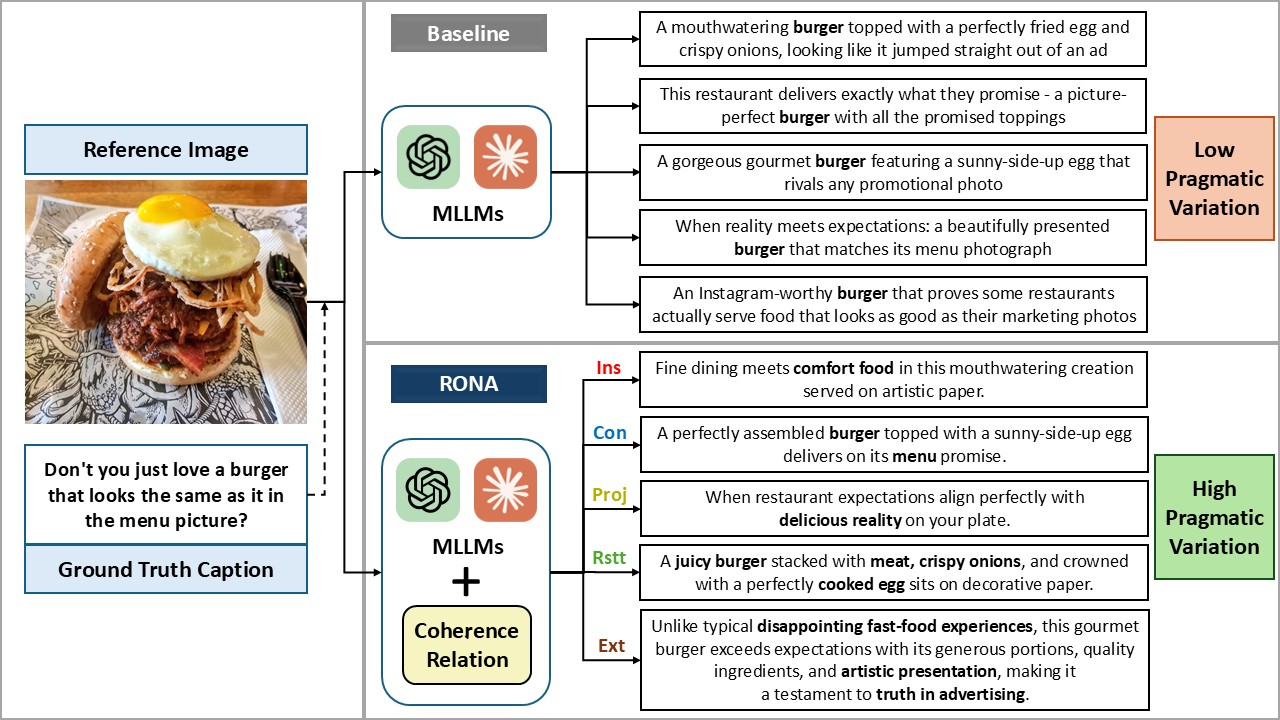"}
    \caption{An overview of {\name}. The CRs used are Insertion (Ins), Concretization (Con), Projection (Proj), Restatement (Rstt) and Extension (Ext). These relations provide a finite but flexible axis of variation for diverse caption generation compared to instruction-based prompts.
    }
    \label{fig:arch_diagram}
\end{figure*}

\section{Introduction}
A {\em Writing Assistant} (WA) is a tool (e.g., Grammarly, Microsoft Copilot, Copy.ai), often powered by Generative AI, that helps users in various writing tasks. WAs have evolved over the years to support users across a multitude of tasks, with AI-powered assistants being adept at generating a wide selection of content. {\em Image Captioning} (\ie generating textual descriptions for given images) is one key task that has seen significant advancements with the introduction of Multi-modal Large Language Models (MLLMs). These pre-trained models have achieved remarkable success in generating captions that accurately describe the visual content of images \cite{Chen2024-yv,Yue2023-ka}. However, real-world image captions across different domains often require more than just a description of the visual elements--\ie they need to convey a central message, provide context, and offer different perspectives on the image \cite{Federico2016-so}. This leads to {\em significant lack of diversity in the generated captions}, limiting the utility of WAs. 

Existing approaches to fostering diversity in image captions have primarily focused on providing a richer vocabulary (\ie syntactic variations) or selecting different components of the image to emphasize (\ie semantic variations) \cite{Bugliarello2021-kq,Li2022-an}. Although these methods have shown promise, they often fail to capture the nuanced ways in which humans communicate through captions as shown in Figure \ref{fig:arch_diagram}. Pragmatic variations, such as multi-modal implicatures and metaphors \cite{Genovesi2020-em}, which utilize meanings or connotations beyond the literal description of the image, are often employed by caption writers to make their messages more engaging and relatable \cite{Weiland2015-zb}. 

To address this challenge, in this work, we propose {\name} (\underline{R}elation-based c\underline{O}here\underline{N}ce-aware c\underline{A}ptioning), a novel prompting strategy for MLLMs inspired by the concept of {\em Coherence Relations} (CRs). Based on the principles of Discourse Theory, CRs provide a structured overview of image-text relationships \cite{Hobbs1978-em,Kress2009-iy,O-Halloran2014-no}, modeling both contextual and pragmatics aspects of language \cite{Ma2025-hv, Mavridou2015-qi}.

We evaluate their effectiveness in image captioning by {\em using CRs as guidelines for generating captions that fulfill specific communicative functions while preserving semantic coherence.}

Our analysis includes popular MLLMs: Claude-3.5 Sonnet V2 \cite{AnthropicUnknown-hu} and GPT-4o \cite{OpenAI2024-hr} on two datasets--\ie news captions (ANNA) and social media captions (Tweet Subtitles). These datasets contain a wide range of visual objects and abstractive captions \cite{Anantha-Ramakrishnan2024-sv}, making this a challenging task for MLLMs. We demonstrate that {\name} outperforms existing baselines on caption diversity while retaining ground truth similarity. Our contributions are as follows:

\begin{itemize}[leftmargin=3.3mm]
    \item We propose {\name}, a novel prompting strategy that leverages Coherence Relations (CRs) to generate pragmatically diverse image captions.
    \item We demonstrate that {\name} outperforms existing MLLM baselines in terms of diversity and ground-truth similarity on news and social media captioning datasets.
    \item Our analysis shows the viability of CRs to be utilized as an axis of variation for Captioning-based writing assistants.
\end{itemize}

\section{Related Work}
\paragraph{Writing Assistants}
MLLMs have enabled WAs to support a variety of writing tasks, with their input ranging from sentence-level suggestions \cite{Gero2022-ws} to long-form writing tasks such as literature reviews and creative writing \cite{Choe2024-gk}, \cite{Singh2023-qs}. In order to improve the Human-AI collaboration experience, there exists a need to incorporate human values into AI-based writing systems and vice versa \cite{Shen2024-zg}, \cite{Lee2024-lf}. Although these studies have focused on broader task domains, specific writing tasks such as caption writing have been less explored \cite{Ng2024-um}, particularly pragmatically diverse captioning which we aim to address.

\paragraph{Image Captioning}
Traditional Captioning models build on task-specific generative architectures to generate faithful and diverse descriptions for images \cite{Mahajan2020-sr,Liu2019-kp}. However, with the introduction of MLLMs, general-purpose models capable of multi-modal representations are utilized for caption generation \cite{Radford2021-ro,Li2023-vq}. To improve the alignment between image-text linkages, Coherence Relations (CR) \cite{Alikhani2020-nr} have been leveraged across different downstream text generation tasks \cite{Alikhani2019-kn,Vempala2019-lh,Sosea2021-hr}. Although popular MLLMs such as GPT-4o and Claude Sonnet 3.5 V2 are poor at predicting and verifying these relationships \cite{Thrush2022-yf,Anantha-Ramakrishnan2025-ce}, existing research does not explore the production capabilities of these models. In our work, we investigate the ability of MLLM-based WAs to utilize CRs as a guidance mechanism for in-context learning.

\section{Methodology}

\paragraph{Coherence Relations}
{\name} leverages in-context explanations of Coherence Relations (CRs) as guidance for generating pragmatically diverse captions. We utilize CRs that characterize both entity-level and scene-level linkages between an image and its expected caption \cite{Xu2022-ie}. Entity-level relations describe the relationships between specific objects in the image and their corresponding elaboration in the caption. Scene-level relations, on the other hand, capture the overall context and narrative of the image, providing a broader understanding of the visual content. The selection of these relations are motivated by their generalizability across different domains. Overall, the 5 types of relations that we utilize are: 

\begin{itemize}[leftmargin=3.3mm]
  \item \textbf{Insertion}: An entity-level relation that describes a type of pragmatic \textit{ellipsis}, \ie where the focal object described in the image and caption does not have an explicit mention in the caption. 
  \item \textbf{Concretization}: An entity-level relation that utilizes an \textit{anchor} object which is prominently referenced in the image and caption, with the caption providing additional meaning about its context.
  \item \textbf{Projection}: An entity-level relation where the caption's description revolves around a particular topic, but this topic is not directly featured in the image. Alternatively, the image contains objects that can be \textit{associated} to this topic instead, forming an implied link between modalities.
  \item \textbf{Restatement}: A scene-level relation that describes the overall context of the image, with the caption providing a more detailed \textit{description} of the visual scene.
  \item \textbf{Extension}: A scene-level relation in which the caption \textit{elaborates} further on the visual scene in terms of new ideas or stories.
\end{itemize}

\begin{table*}[!ht]
    \centering
    \small
        \begin{tabularx}{\linewidth}{@{} l | c | YYYY @{}}
        \toprule

        \textbf{Task} & \textbf{Model} & \textbf{BLEURT ↑} & \textbf{CLIPScore ↑} & \textbf{Self-BLEURT ↓} & \textbf{Div-2 ↑} \\
        \midrule

        \multirow{2}{*}{Image-only} & Claude & -1.227 & 14.049 & 0.226 & 0.860 \\
         & {\name} + Claude & \textbf{-1.141} & \textbf{14.068} & \textbf{0.108} & \textbf{0.903} \\

        \midrule
        
        \multirow{2}{*}{Image-only} & GPT-4o & -1.237 & 13.117 & \textbf{0.198} & \textbf{0.885} \\
        & {\name} + GPT-4o & \textbf{-1.137} & \textbf{14.505} & 0.205 & 0.879 \\
        
        \midrule
        \midrule

        \multirow{2}{*}{Image + Caption} & Claude & -0.931 & 13.833 & 0.294 & 0.843 \\
         & {\name} + Claude & \textbf{-0.879} & \textbf{13.866} & \textbf{0.158} & \textbf{0.882} \\

        \midrule

        \multirow{2}{*}{Image + Caption} & GPT-4o & -0.650 & 13.200 & \textbf{0.355} & 0.805 \\
        & {\name} + GPT-4o & \textbf{-0.615} & \textbf{13.891} & 0.383 & \textbf{0.823} \\

        \bottomrule

        \end{tabularx}
       \caption{Results for Diverse Image Captioning with {\name} on the Tweet Subtitles Dataset.
       }
        \label{table:metrics_tweets}
\end{table*}

\begin{table*}[!ht]
    \centering
    \small
        \begin{tabularx}{\linewidth}{@{} l | c | YYYY @{}}
        \toprule

        \textbf{Task} & \textbf{Model} & \textbf{BLEURT ↑} & \textbf{CLIPScore ↑} & \textbf{Self-BLEURT ↓} & \textbf{Div-2 ↑} \\
        \midrule

        \multirow{2}{*}{Image-only} & Claude & -1.191 & \textbf{14.617} & 0.258 & 0.854 \\
         & {\name} + Claude & \textbf{-1.038} & 14.471 & \textbf{0.134} & \textbf{0.899} \\

        \midrule
        
        \multirow{2}{*}{Image-only} & GPT-4o & -1.159 & 13.954 & 0.249 & \textbf{0.883} \\
        & {\name} + GPT-4o & \textbf{-1.057} & \textbf{15.022} & \textbf{0.209} & 0.878 \\
        
        \midrule
        \midrule

        \multirow{2}{*}{Image + Caption} & Claude & -0.669 & \textbf{14.582} & 0.341 & 0.845 \\
         & {\name} + Claude & \textbf{-0.559} & 14.549 & \textbf{0.217} & \textbf{0.883} \\

        \midrule

        \multirow{2}{*}{Image + Caption} & GPT-4o & \textbf{-0.356} & 14.338 & 0.436 & 0.796 \\
        & {\name} + GPT-4o & -0.363 & \textbf{14.869} & \textbf{0.394} & \textbf{0.824} \\

        \bottomrule

        \end{tabularx}
       \caption{Results for Diverse Image Captioning with {\name} on the ANNA Dataset.}
        \label{table:metrics_anna}
\end{table*}

\paragraph{Datasets} \label{datasets}
Popular datasets such as COCO Captions \cite{Chen2015-qj} or Flickr30K \cite{Young2014-sr} are often used for image captioning evaluation, but the ground-truth captions do not cater to sharing messages or perspectives that are more aligned with human-written captions. Instead, we select datasets from 2 different task domains for image captioning: news and social media. These domains provide representative examples of real-world scenarios for the usage of WAs: (1) The Tweet Subtitles dataset \cite{Xu2022-ie} contains 16,000 image-text pairs sourced from Twitter and cleaned to remove noisy, low-quality samples, and  (2) ANNA \cite{Anantha-Ramakrishnan2024-sv} on the other hand contains 29,625 image-text pairs collected from The New York Times news articles focusing on non Named Entity objects. Both datasets contain ``abstractive" or non-descriptive captions with a wide range of image subjects and topics. For our evaluation, we used the entire test set of 1,600 samples from Tweet Subtitles and a random sample of 1,500 images from the test set of ANNA.

\section{Experiments}
\paragraph{Task Types} \label{task-types}
For our analysis of the effectiveness of {\name}, we define 2 task types: \textit{Image-only} and \textit{Image + Caption}. In the \textit{Image-only} task, we define this as a classic image captioning task in which the model is provided only with the image as input. On the other hand, for the \textit{Image + Caption} task, we provide the model with both the image and a ground-truth caption as input. Since both of these components are part of understanding the overall meaning of an image-caption pair, we wish to understand how MLLMs utilize both modalities to generate diverse captions without the divergence of meaning. This is similar to the prompt-guided image captioning task for MLLMs \cite{Hu2023-ue}. In both tasks, the baseline MLLM is prompted to use the inputs to generate captions with ``as much diversity as possible while retaining their original meaning and message." {\name} utilizes in-context learning where simplified definitions of CRs are provided as system prompts. We generate 5 captions per input for each type of task, with {\name} generating one caption per CR. Additional generated caption examples are presented in Appendix Section \ref{appendix:additional_examples}.

\paragraph{Evaluation Metrics}
To evaluate the performance of MLLMs on the task of diverse captioning, we measure 4 key attributes: image-caption similarity, ground truth caption similarity, contextual diversity, and bi-gram diversity. CLIPScore \cite{Hessel2021-we} effectively measures image-caption similarity by converting both modalities into a common representation space. For validating similarity of generated captions with the ground truth text, we turn to BLEURT \cite{Sellam2020-wy} score. Unlike traditional similarity metrics such as BLEU \cite{Papineni2001-gb}, METEOR \cite{Lavie2007-ts} and BERTScore \cite{Zhang*2020-xq}, BLEURT is trained to balance contextual similarity and human preference judgments, making it better suited for non-descriptive captions. All similarity metrics are computed pairwise between the ground truth modality and generated captions, with the average score reported in our benchmarks. For judging contextual diversity, we reformulate it as a task of minimizing the pairwise similarity between generated captions. This homogenization process is applied to BLEURT score, converting it into the diversity metric Self-BLEURT \cite{Shaib2024-xc}. Finally, we calculate the overall bi-gram diversity of generated captions using the Div-2 metric \cite{Shetty2017-ld}, which reports the ratio of unique bi-grams to the total count of bi-grams in a sentence.

\section{Results}
\paragraph{{\name} Improves Relevance and Diversity}

We present our evaluation of MLLMs on the task of Diverse Image Captioning in Tables \ref{table:metrics_tweets} and \ref{table:metrics_anna}. Our assessment spans 8 different settings: 2 tasks per dataset, 2 models per task and 2 different dataset domains. From our results, both GPT-4o and Claude combined with {\name} outperforms their respective baselines in 7/8 settings on both ground truth similarity and diversity metrics. Particularly, we see a positive agreement between Div-2 and Self-BLEURT, as they rate captions from {\name}-based models as more diverse over 5/8 baselines. With image \& text similarity metrics such as BLEURT and CLIPScore preferring {\name}-based models 7/8 and 6/8 times over baselines respectively, we can conclude that our observed diversity has not come at the cost of contextual relevance.

\paragraph{Diversity and Similarity Trade-off Across Modalities}
From our experiments across task types, we observe a small decrease in image similarity but improved caption similarity and diversity in the Image + Caption task compared to Image-only task. This confirms that image-only descriptive captioning approaches are limited in terms of expression and rely heavily on listing visual features. This motivates the need for WAs to be evaluated on captions with pragmatic variations to test their true ability in understanding the overall message of a sample.

\section{Conclusions}
We propose {\name}, a Coherence Relation-based prompting strategy, providing a framework for expressive and diverse image caption generation. Our study presents a holistic evaluation of top MLLMs on their ability to utilize these relationships through in-context learning. Our results show that {\name} enables the generation of a greater variety of captions while improving their overall semantic and contextual relevance across domains. {\name} serves as a new baseline for future work leveraging image-text relationships for improving the quality of Multi-modal Writing Assistants.

\clearpage
\section*{Limitations}
Our current analysis of {\name} is limited to a couple of top-performing MLLM architectures. Evaluating how open-source MLLMs can leverage CRs for image captioning is a part of our future work. Additionally, our evaluation strategy does not validate the prompt following accuracy of MLLMs in adhering to specific CRs, which would be a significant challenge for smaller, low-resource models. These inaccuracies may lead to hallucinations, harming the factual accuracy of generated captions. Incorporating Factual Consistency metrics and Human preference ratings to identify potential types of hallucinations in diverse captioning tasks is a direction of future work we wish to pursue.

\section*{Ethics Statement}
We acknowledge the potential for alternate prompting strategies like {\name} to be used for generating misleading content, especially from specific domains such as news media. However, from our evaluation, we find that MLLM safety filters are robust in capturing potentially harmful content in either the input images or captions as described in Appendix Section \ref{appendix:post_processing}. With CRs leveraging pragmatic and common-sense knowledge of MLLMs to generate diverse captions, there exists a possibility of model biases and stereotypes clouding the quality of our generations. This is especially a problem in cases where culturally sensitive material is present in our input samples. We advocate for the responsible use of Writing Assistants with adequate human oversight to prevent such situations.

\section*{Acknowledgments}
This research was in part supported by the U.S. National Science Foundation (NSF) award \#1820609. Part of the research results were obtained using the computational resources provided by CloudBank (\url{https://www.cloudbank.org/}), which was supported by the NSF award \#1925001.

\bibliography{acl_latex}

\clearpage
\appendix
\section*{Appendix}
\label{sec:appendix}

\section{Data Preparation}
\label{appendix-data-prep}

This section sheds light on the methods used while preparing all the datasets mentioned in this paper for model evaluation. We verify both datasets used to evaluate {\name} have a permissive license that allows usage for research purposes (Tweet Subtitles: \textbf{MIT License} and ANNA: \textbf{CC BY-NC-SA 4.0 License}).

\subsection{Tweet Subtitles}
This dataset contains two types of captions for tweets: actual and text generated by an image captioning model. We use only the \textbf{actual} caption as part of our evaluation.

\subsection{ANNA}
As mentioned in Section \ref{datasets}, we randomly sample \textbf{1,500 examples} from ANNA to construct our test set. We use a random seed of $42$ to ensure reproducibility.

\section{Proprietary Model Details}

\paragraph{OpenAI GPT:} We access the GPT-4o model via a custom deployment using Azure OpenAI. We evaluate \texttt{gpt-4o-2024-11-20} with a custom safety filter to restrict content of \textit{high} severity level.

\paragraph{Anthropic Claude:} We access Claude 3.5 Sonnet v2 via the Vertex AI API, using Google Cloud. We evaluate \texttt{claude-3-5-sonnet-v2@20241022}.

\section{Prompt Templates}
We use two types of prompt templates, both with different system/user messages for the evaluation of tasks mentioned in Section \ref{task-types}. The different prompts and system messages used are present in the appendix.

\section{Postprocessing MLLM Responses}
\label{appendix:post_processing}
Since both datasets feature image-caption pairs on a wide selection of topics, around \textbf{50 images} were flagged by the safety filter or rejected by the model for captioning. To ensure test set consistency, we remove these examples from our evaluation.

\section{Example Captions from {\name}}
\label{appendix:additional_examples}
We also provide a few examples of captions generated in each setting below. All captions were generated using Claude 3.5 Sonnet v2.

\subsection{ANNA Examples}
\begin{figure}[ht]
    \centering
    \includegraphics[width=0.7\linewidth]{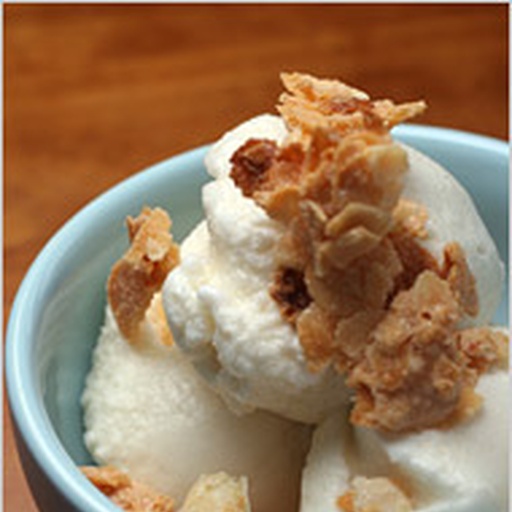}
    \caption*{\textbf{Caption:} JUST NUTS: The creaminess of this dessert comes from churned almond milk.}
    \label{fig:anna-example}
\end{figure}

\paragraph{Baseline (Image-only)}
\begin{itemize}
    \item Creamy vanilla ice cream topped with crispy honeycomb pieces.
    \item Homemade ice cream dessert garnished with crunchy toffee bits.
    \item Sweet and crunchy: vanilla gelato with caramelized sugar shards.
    \item Classic vanilla scoops adorned with brittle candy fragments.
    \item Artisanal ice cream featuring golden honeycomb crumble topping.
\end{itemize}

\paragraph{{\name} (Image-only)}
\begin{itemize}
    \item \textbf{Insertion:} The perfect way to end a summer evening on the patio.
    \item \textbf{Concretization:} Vanilla ice cream garnished with crispy honeycomb pieces in a blue ceramic bowl.
    \item \textbf{Projection:} Sweet memories of childhood desserts come flooding back with every spoonful.
    \item \textbf{Restatement:} Scoops of creamy white ice cream topped with crunchy golden candy pieces, served in a light blue bowl against a wooden background.
    \item \textbf{Extension:} This artisanal dessert combines locally-sourced dairy with house-made honeycomb brittle, showcasing how simple ingredients can create an extraordinary taste experience when crafted with care.
\end{itemize}

\paragraph{Baseline (Image + Caption)}
\begin{itemize}
    \item Creamy almond milk ice cream topped with crunchy toasted nut brittle.
    \item Dairy-free frozen dessert made with almond milk and garnished with caramelized nuts.
    \item Smooth vegan ice cream featuring almond milk, decorated with crispy nut pieces.
    \item Artisanal almond milk gelato adorned with handmade nutty crumble.
    \item Plant-based frozen treat with rich almond flavor and crunchy nut topping.
\end{itemize}

\paragraph{{\name} (Image + Caption)}
\begin{itemize}
    \item \textbf{Insertion:} A refreshing frozen treat perfect for hot summer days, churned to creamy perfection.
    \item \textbf{Concretization:} Scoops of almond milk ice cream topped with crunchy toasted nuts in a blue ceramic bowl.
    \item \textbf{Projection:} The dairy-free dessert revolution takes shape in this plant-based frozen delight.
    \item \textbf{Restatement:} Creamy white almond milk ice cream served in a light blue bowl, garnished with crushed nuts on top.
    \item \textbf{Extension:} This artisanal vegan ice cream demonstrates how nut-based alternatives can rival traditional dairy desserts, offering both sustainable ingredients and decadent taste while catering to those with dietary restrictions.
\end{itemize}

\subsection{Tweet Subtitles Examples}
\begin{figure}[ht]
    \centering
    \includegraphics[width=0.8\linewidth]{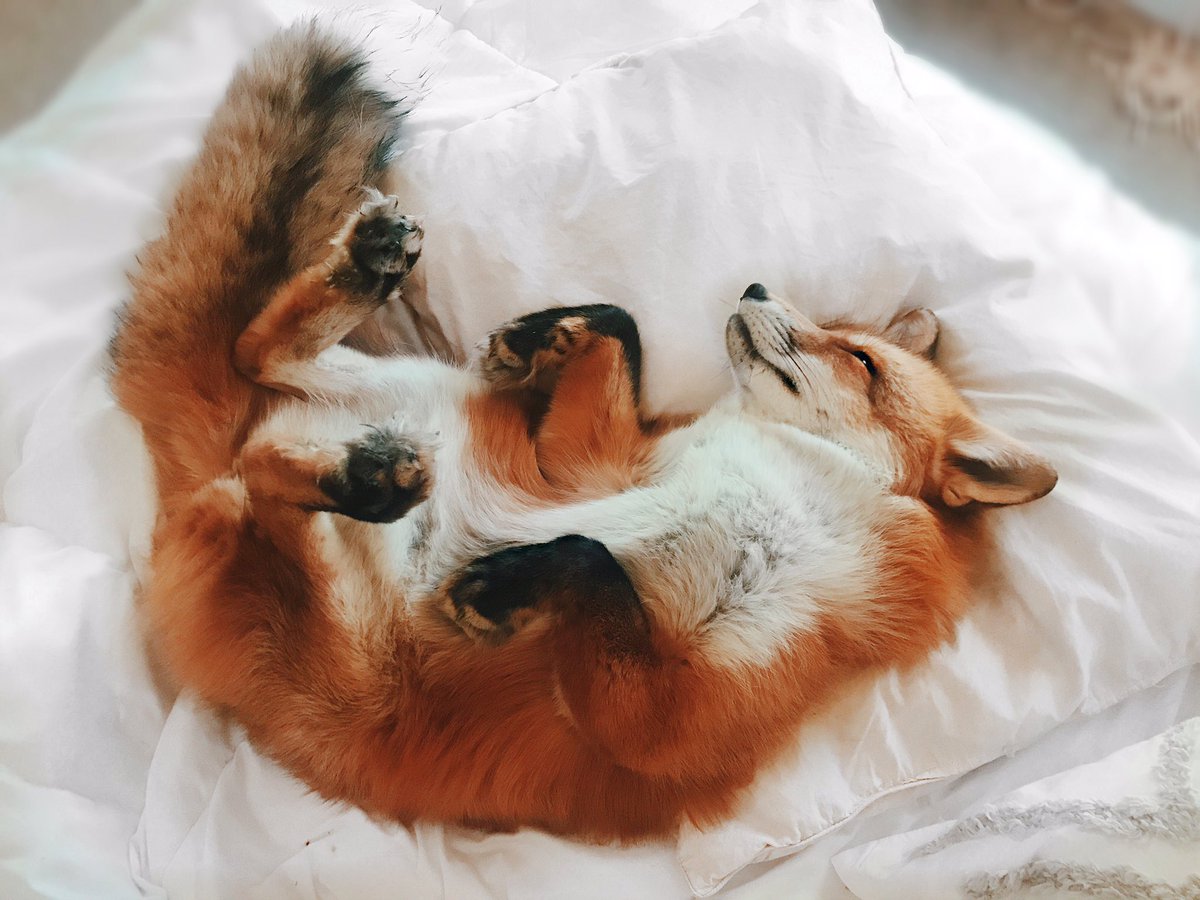}
    \caption*{\textbf{Caption:} It's raining, it's pouring. The sleepy fox is snoring.}
    \label{fig:tweets-example}
\end{figure}

\paragraph{Baseline (Image-only)}
\begin{itemize}
    \item Sleepy fox curled up contentedly on white sheets.
    \item Red fox showing its fluffy belly while napping peacefully.
    \item A comfortable fox enjoying a lazy day in bed.
    \item Beautiful russet-colored fox in complete relaxation mode.
    \item Wild spirit at rest: fox taking an adorable snooze on soft bedding.
\end{itemize}

\paragraph{{\name} (Image-only)}
\begin{itemize}
    \item \textbf{Insertion:} Sweet dreams are made of soft white sheets and peaceful afternoon naps.
    \item \textbf{Concretization:} A red fox lies contentedly on its back, nestled in crisp white bedding.
    \item \textbf{Projection:} Finding complete comfort and trust in one's surroundings is a rare and precious thing.
    \item \textbf{Restatement:} A fox is sleeping upside down on white sheets, with its red-orange fur contrasting against the white background, showing its belly and paws in the air.
    \item \textbf{Extension:} Like many domestic pets, this rescued fox has adapted to human comforts, choosing the soft embrace of fresh linens over its natural forest bedding, showing how wild animals can adapt to domesticated environments when given the chance.
\end{itemize}

\paragraph{Baseline (Image + Caption)}
\begin{itemize}
    \item A cozy fox napping peacefully on white sheets.
    \item Red fox curled up in blissful slumber, paws in the air.
    \item Playful fox showing its belly while taking a comfortable nap.
    \item Beautiful russet-colored fox in deep sleep on soft bedding.
    \item Adorable fox stretched out and snoring on pristine white linens.
\end{itemize}

\paragraph{{\name} (Image + Caption)}
\begin{itemize}
    \item \textbf{Insertion:} The gentle pitter-patter outside creates the perfect atmosphere for afternoon rest.
    \item \textbf{Concretization:} A sleepy red fox lies curled up, snoring peacefully on white bedding.
    \item \textbf{Projection:} Dreaming of chasing rabbits while the storm rages outside.
    \item \textbf{Restatement:} A red fox is sleeping on its back, completely relaxed on white sheets while it's raining outside.
    \item \textbf{Extension:} As rain drums against the windowpane, this rescued fox demonstrates how domesticated wildlife can find comfort in human spaces, sprawled out on clean linens just like a household pet.
\end{itemize}

\begin{figure*}[t]
    \centering
    \begin{tcolorbox}[title={System Message for Baseline Captioning}, after skip=0pt, boxsep=5pt, width=\linewidth]

    You are an expert linguist, and your task is to write image captions.
    
    \end{tcolorbox}
\end{figure*}

\begin{figure*}[t]
    \centering
    \begin{tcolorbox}[title={System Message for {\name}}, colframe = blue!30, colback = blue!10, coltitle = blue!20!black, after skip=0pt, boxsep=5pt, width=\textwidth]

    You are an expert linguist, and your task is to write image captions with the help of Coherence Relations. A coherence relation describes the structural, logical, and purposeful relationships between an image and its caption, capturing the author's intent. \\
    
    These are the possible coherence relations you can assign to an image-text pair:
    
    - Insertion: The salient object described in the image is not explicitly mentioned in the text. \\
    - Concretization: Both the text and image contain a mention of the main visual entity. \\
    - Projection: The main entity mentioned in the text is implicitly related to the visual objects present in the image. \\
    - Restatement: The text directly describes the image contents. \\
    - Extension: The image expands upon the story or idea in the text, presenting new elements or elaborations, effectively filling in narrative gaps left by the text.
    
    \end{tcolorbox}
\end{figure*}

\begin{figure*}[t]
    \centering
    \begin{tcolorbox}[title={Prompt for Baseline Captioning}, , colframe = orange!30, colback = orange!10, coltitle = orange!20!black, after skip=0pt, boxsep=5pt, width=\textwidth]
    
    \textbf{System} \\
    <insert-system-message> \\
    
    \textbf{User} \\
    You will be given an \textcolor{red}{image} (or) \textcolor{red}{image-caption pair} as input. Analyze the image and write 5 suitable captions that are diverse, but relevant. Create diverse captions while retaining the same overall meaning of the original image-caption pair. \\

    Return the captions as a JSON Array with the following format: \\
    \texttt{
         [\\"<insert-caption-text-1>", \\"<insert-caption-text-2>",\\ "<insert-caption-text-3>",\\ "<insert-caption-text-4>",\\ "<insert-caption-text-5>"\\]
    }
    \\
    
    \textbf{<insert-image>} \\
    \textbf{<insert-caption>} \\
    
    \end{tcolorbox}
\end{figure*}

\begin{figure*}[t]
    \centering
    \begin{tcolorbox}[title={Prompt for {\name}}, colframe = red!30, colback = red!10, coltitle = red!20!black, after skip=0pt, boxsep=5pt, width=\textwidth]
    
    \textbf{System} \\
    <insert-system-message> \\
    
    \textbf{User} \\
    You will be given an \textcolor{red}{image} (or) \textcolor{red}{image-caption pair} as input. Write 5 image captions, one for each coherence relation as your output. \\

    Return the captions as a JSON object with the following format: \\
    \texttt{
        \{ \\
        "Insertion": "<insert-caption-text-1>", \\ "Concretization": "<insert-caption-text-2>", \\ "Projection": "<insert-caption-text-3>", \\ "Restatement": "<insert-caption-text-4>", \\ "Extension": "<insert-caption-text-5>" \\
        \}
    }
    \\
    
    \textbf{<insert-image>} \\
    \textbf{<insert-caption>} \\
    
    \end{tcolorbox}
\end{figure*}

\clearpage

\end{document}